\documentclass[conference]{IEEEtran}
\usepackage{times}

\usepackage{comment}
\usepackage[numbers]{natbib}
\usepackage{multicol}
\usepackage[bookmarks=true]{hyperref}
\usepackage{amsmath}
\usepackage{amssymb}
\usepackage{graphicx}
\usepackage{tabu}
\usepackage{adjustbox,multirow}
\usepackage{makecell}
\usepackage{caption}
\usepackage{subcaption}
\usepackage[ruled,vlined]{algorithm2e}
\SetKwInput{KwInput}{Input}                
\SetKwInput{KwOutput}{Output}              
\usepackage{algorithmicx,algpseudocode,float}
\usepackage{pgfplots}
\pgfplotsset{compat=1.9}

\newcommand{\floor}[1]{\lfloor #1 \rfloor}

  {
      
      \newtheorem{property}{Property}
  }

\usepackage{longtable}
\usepackage{makecell, cellspace, caption}
\usepackage{array}
\newcolumntype{L}[1]{>{\raggedright\let\newline\\\arraybackslash\hspace{0pt}}m{#1}}
\newcolumntype{C}[1]{>{\centering\let\newline\\\arraybackslash\hspace{0pt}}m{#1}}
\newcolumntype{R}[1]{>{\raggedleft\let\newline\\\arraybackslash\hspace{0pt}}m{#1}}

\begin{document}

\title{



Extended Tree Search for Robot Task and Motion Planning


}




%


\author{\authorblockN{Tianyu Ren\authorrefmark{1},
Georgia Chalvatzaki\authorrefmark{1},
and Jan Peters\authorrefmark{1}}
\authorblockA{\authorrefmark{1}Computer Science Department, Technische Universität Darmstadt}}

\maketitle

\begin{abstract}
Integrated task and motion planning (TAMP) is desirable for generalized autonomy robots but it is challenging at the same time. TAMP requires the planner to not only search in both the large symbolic task space and the high-dimension motion space but also deal with the infeasible task actions due to its intrinsic hierarchical process.
We propose a novel decision-making framework for TAMP by constructing an extended decision tree for both symbolic task planning and high-dimension motion variable binding. We integrate top-k planning for generating explicitly a skeleton space where a variety of candidate skeleton plans are at disposal. Moreover, we effectively combine this skeleton space with the resultant motion variable spaces into a single \textit{extended} decision space. Accordingly, we use Monte-Carlo Tree Search (MCTS) to ensure an exploration-exploitation balance at each decision node and optimize globally to produce optimal solutions. The proposed seamless combination of symbolic top-k planning with streams, with the proved optimality of MCTS, leads to a powerful planning algorithm that can handle the combinatorial complexity of long-horizon manipulation tasks. We empirically evaluate our proposed algorithm in challenging robot tasks with different domains that require multi-stage decisions and show how our method can overcome the large task space and motion space through its effective tree search compared to its most competitive baseline method.

\end{abstract}

\IEEEpeerreviewmaketitle

\section{Introduction}
\begin{figure}
    \centering
    \begin{subfigure}[b]{0.49\textwidth}
        \centering
        \includegraphics[width=0.9\textwidth]{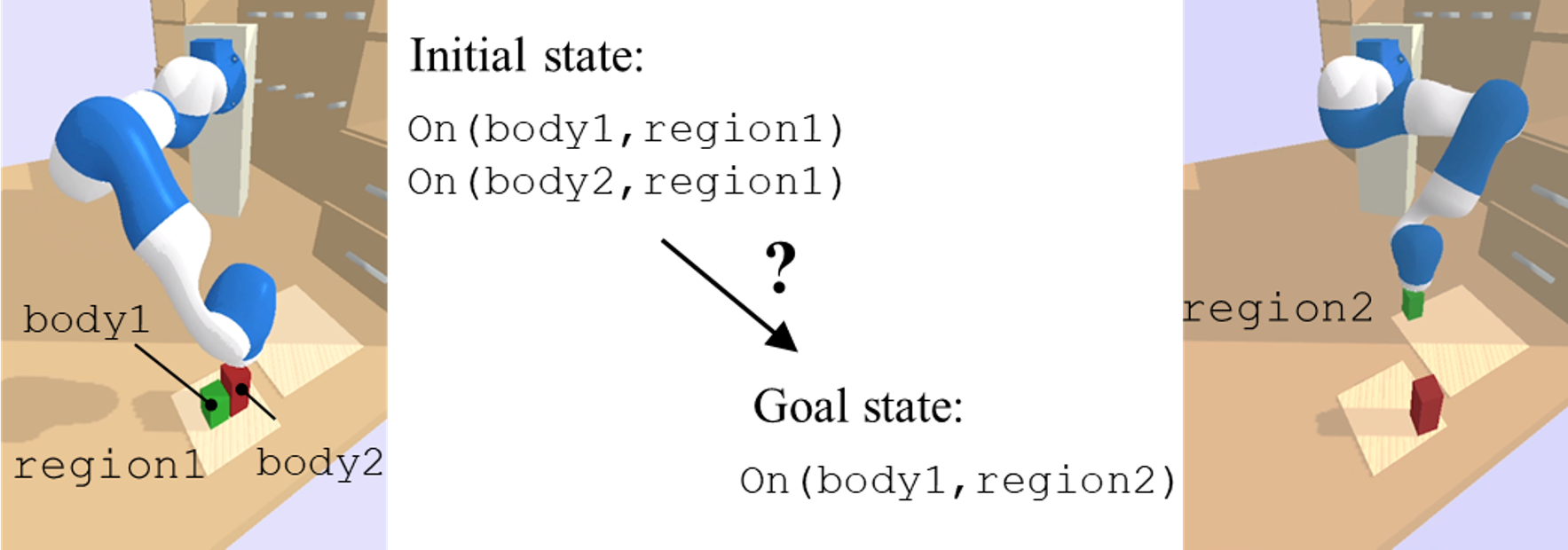}
            \vspace{-0.18cm}
        \caption{\small The Unpacking problem}
        \label{fig:toy_problem}
    \end{subfigure}
    \vfill
    \begin{subfigure}[b]{0.49\textwidth}
        \centering
        \includegraphics[width=0.9\textwidth]{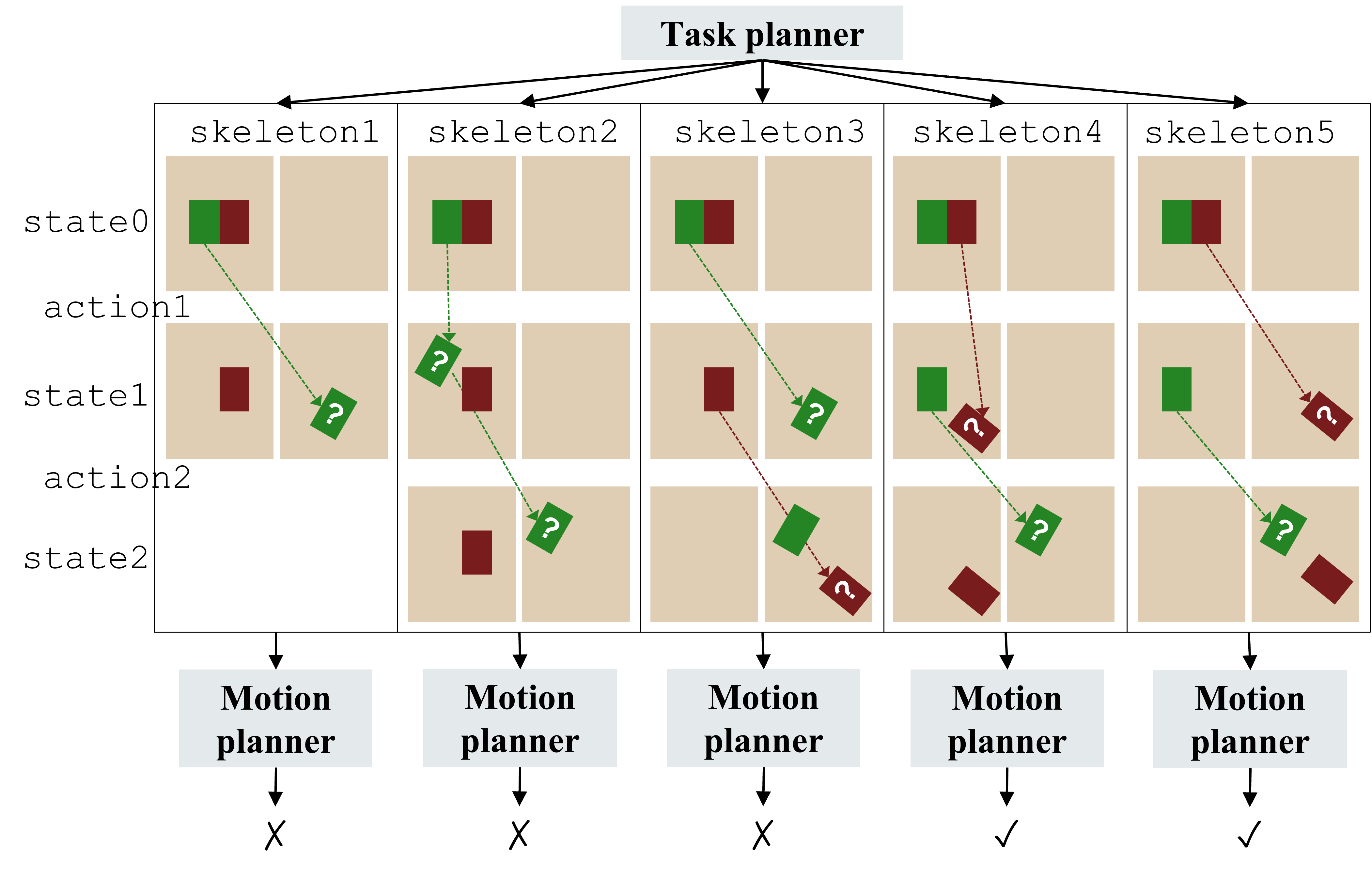}
            \vspace{-0.18cm}
        \caption{\small A hierarchical (or "search-then-sample" \cite{garrett2021integrated}) TAMP workflow}
        \label{fig:candidate_sk}
    \end{subfigure}
        \vspace{-0.45cm}
    \caption{\small The unpacking problem. The taller red body in the environment must be relocated before the green one can be reached without collision.}
    \label{fig:1}
\end{figure}

 Traditional approaches like \cite{fikes1994strips} considered the long-horizon manipulation as a strict decomposition of symbolic sub-tasks to be solved by motion planners independently. However, in most cases dependencies are unneglectable between symbolic and geometric levels. Consider a toy problem (Fig\,\ref{fig:toy_problem}) with its TAMP solutions  (Fig\,\ref{fig:toy_problem}), it is obvious that the success of the motion planners depends on the outputs of the task planner. On the other hand, the task planner should be aware of how its outputs are evaluated by motion planners to avoid spending too much time on infeasible skeletons. Therefore, planning techniques are required to integrate both task planning and motion planning. We are interested in an integrated planner that is able to search in both a large action space and a large motion space while addressing the infeasible skeleton issue resulting from the combination of the two spaces.
 
\textit{\underline{Large task spaces.}}
In comparison to multi-modal motion planning (MMMP) \cite{hauser2010multi,barry2013hierarchical} that mainly extends the classic motion planning (for collision avoidance) to include changing the state of other objects in the domain, a TAMP problem features large abstract state spaces with additional state variables that have no geometric meaning, such as whether the food is cleaned in a cooking task \cite{garrett2021integrated}. To make this large action space tractable, the communities usually take relational representations such as PDDL \cite{mcdermott1998pddl} to compactly describe the planning domain by exploiting its underlying regularities \cite{garrett2018stripstream,garrett2020pddlstream,dantam2016incremental}.
Thus, task planning in TAMP are generally more complex and universal than in MMMP, and they require substantial search effort with dedicated AI planners.

\textit{\underline{Large motion spaces.}}
A skeleton must have its symbolic actions grounded by concrete motion values (\textit{bindings}) so that it can be executed by physical robots. Therefore, low-level motion planning with high-dimension variables are ultimately unavoidable. The motion space depends on both the parametrization of the abstract actions and the length of the skeleton. For example in Fig.\,\ref{fig:candidate_sk}, we have $\langle x_1,y_1,w_1 \rangle \in \mathbb{R}^3$ to search for the $x$- and $y$-position of \texttt{body1} and its orientation $w_1$ in \texttt{skeleton1}. While in \texttt{skeleton2}, the dimension of the motion space gets doubled: $\langle x_1,y_1,w_1,x_2,y_2,w_2 \rangle \in \mathbb{R}^6$. Since even larger motion spaces in robot manipulation are expected, an efficient motion planner is highly desired.

\textit{\underline{Infeasible task skeletons.}} 
The state abstraction of TAMP inevitably prevents the task planners from accessing the detailed features of the environment that are possibly vital to the problem's solution, For example in Fig\,\ref{fig:toy_problem}, since the task planner receives no information from the symbolic initial state about the blockade of the taller body, thus it will probably generate \texttt{skeleton1} as the first attempt.
In AI planning research, this issue is commonly described as an \textit{incomplete domain description} \cite{weber2011planning,zhuo2013refining} in the sense that the task planner does not have enough domain knowledge to generate absolutely correct results.
To recover from such a deal end, the task planner in TAMP must provide alternative skeletons. Further, it must optimally select skeletons for further examination based on the binding search report from motion planning.

This paper proposes extended tree search for TAMP (\textit{eTAMP}) that is able to address the above challenges in a single framework with a simple pipeline~\footnote{The code is online available at \href{https://github.com/ttianyuren/eTAMP}{https://github.com/ttianyuren/eTAMP}}. In eTAMP, the first step (Sec.\,\ref{sec:topk-sk}) is to solve the task planning problem with a top-k planner and acquire diverse high-level plans. A skeleton space is build on these diverse plans for the following skeleton selection. In the second step (Sec.\,\ref{sec:extended-tree-search}), a decision tree is built with its first decision node for skeleton selection. From the second decision nodes on, a motion planner start binding search for each skeleton branch. After a preset number of roll-outs, eTAMP takes the tree branch of the highest value as the output plan. We summarize the eTAMP algorithm in Sec.\,\ref{sec:etamp}.
The main contribution of this work is twofold:
\begin{itemize}
    \item we propose a top-k skeleton planner based on generic top-k algorithms that generates diverse high-level plans in large task spaces;
    \item we propose an extended decision tree that models the skeleton selection under infeasible task skeletons and binding search in large motion spaces within the same optimal decision process.
\end{itemize}

We evaluate the proposed method on challenging robotic tasks defined in various domains, using both a 7-degrees-of-freedom (dof) robot and a 10-dof mobile manipulator. Our empirical results prove the ability of eTAMP to find feasible plans under various difficulties. 

\section{Related Work}

Broadly speaking, TAMP contains also multi-modal motion planning problems that have been frequently approached as an optimization problem using logic geometric programming \cite{toussaint2015logic}, or multi-modal motion planning with motion-modes switches for various tasks \cite{kingston2020informing,mordatch2012discovery,mordatch2012contact}. However, these methods are designed for particular robot domains, e.g., desktop manipulation \cite{toussaint2015logic}, in-hand manipulation \cite{mordatch2012contact} and humanoid locomotion \cite{mordatch2012discovery}. They are applicable to neither problems with large action space such as the game of Hanoi Tower nor those with non-geometric actions. On the other hand, general-purpose TAMP systems generically use symbolic AI planners for  efficient problem solving in task spaces, where the relational language PDDL is used for the description of the symbolic domain and problem. A comprehensive review over integrated TAMP approaches can be found in \cite{garrett2021integrated}. Here we discuss only those that are closely
related with our method.

In in the earliest approaches such as \textit{aSyMov} \cite{cambon2009hybrid} and \textit{SMAP} \cite{plaku2010sampling}, the TAMP framework is more of a single-direction flow from task planning to motion planning. Based on the established links between symbolic descriptions and their geometric counterpart, high-dimension motion planning can be effectively guided by symbolic action plans. Nevertheless, these TAMP systems become less efficient when the problems are insufficiently constrained by symbolic descriptions and infeasible skeletons arise. By allowing action planners to validate preconditions of the
symbolic actions in the geometric space when computing action plans, TAMP methods such as \textit{FFRob} \cite{garrett2018ffrob} and \textit{PDDLStream} \cite{garrett2020pddlstream,garrett2018stripstream} strengthen the interplay between action planning and motion planning. These methods, however, require users to craft carefully the symbolic description that includes specifications of the geometric constraints in all high-level actions.

Without the explicit specification of geometric constraints, the TAMP system will expect more infeasible skeletons and a large motion space to search. Since the feasibility of a skeleton is ultimately evaluated via motion planning, the task planner must be able to take feedback from the motion planning results and react accordingly. 
Approaches as proposed in \cite{srivastava2014combined}, \cite{dantam2018incremental} and \cite{shah2020anytime} incrementally refine the high-level representations using the
feedback from the low-level planning while searching for the motion bindings. Failures in motion planning are symbolized into logical predicates that are used to update the high-level models. Then new skeletons can be generated based on these updated models and they could possibly bypass those geometric failures. However, symbolization and expression of the geometric failures in high-level models are difficult when the low-level decisions are continuous. Therefore, it is a common practice for these methods to discretize some dimensions of the motion spaces that are considered to potentially contribute to the high-level models. This partial discretization of the motion space demands human expertise and makes the completeness of their planners questionable. 

In this work, we build on current advances in TAMP and generic AI planning.
To integrate search in both large task spaces and large motion spaces while minimising human inputs, we explore a new method in this study. 
Unlike the above-mentioned TAMP approaches, our method explicitly models the skeleton selection process. An optimal skeleton selection is achieved by (i) maintaining an explicit skeleton space and (ii) evaluating consistently the value of each skeleton. 

We construct a skeleton space with various high-level plans by using a generic top-k planner as the task planner.
Top-k planning is one way of obtaining such a set, by finding a group of diverse solutions of size $k$. With a variety of candidate skeletons, it is possible to make a detour to avoid infeasible skeletons. Based on \textit{FastDownward} \cite{helmert2006fast,torralba2014symba}, \cite{speck2020symbolic} presents a complete top-k planner \textit{SYM-K} that scales efficiently to large sizes of $k$. 

For value estimation of each skeleton, we rely on the back-propagation mechanism of MCTS from the binding search of this skeleton.
Specifically, we use \textit{PW-UCT} (Upper Confidence bounds for Trees with Progressive Widening) \cite{coulom2007computing,auger2013continuous,chaslot2008progressive}, to solve the sequential binding problem over large motion spaces.

\section{Preliminaries}

\begin{figure}
    \centering
    \includegraphics[width=0.49\textwidth]{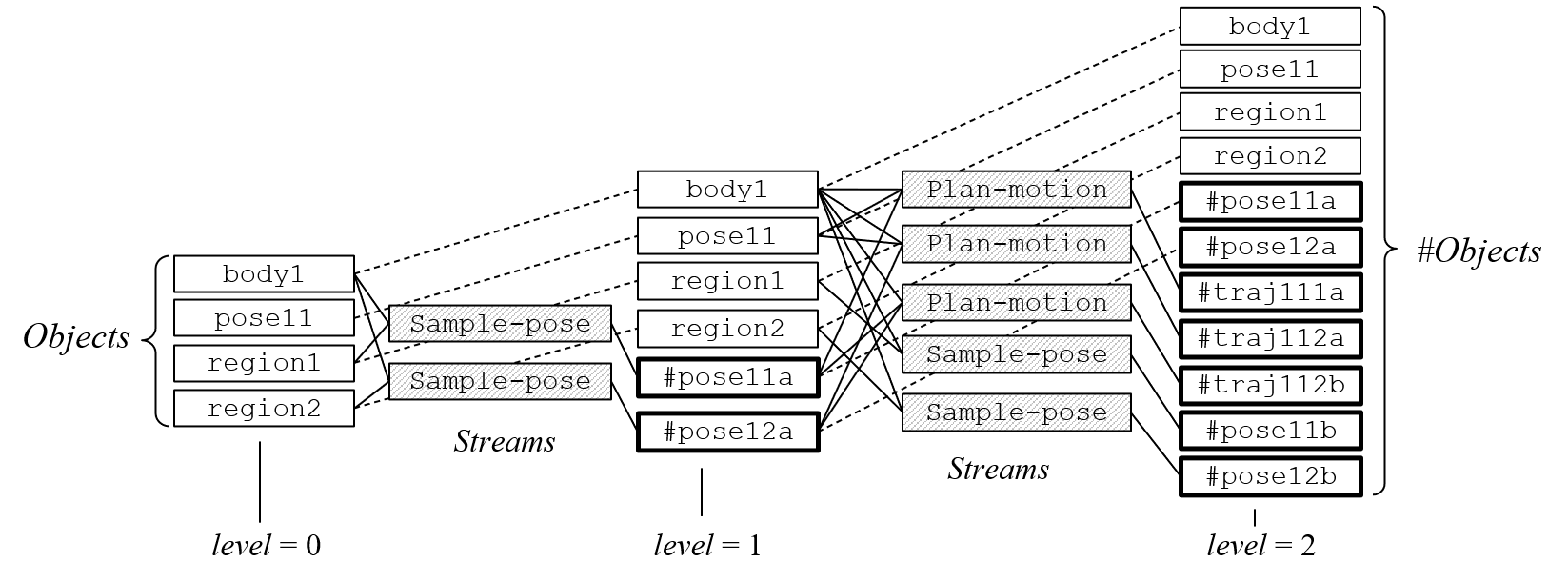}
    \caption{\small An example implementation of OPTMS-EXPD. At each level, streams whose precondition is satisfied generate new optimistic objects. The enrichment of the state is not shown in this figure.}
    \label{fig:enrich}
    \vspace{-0.5cm}
\end{figure}

\noindent\textit{\underline{PDDLStream description.}} 
PDDLStream \citep{garrett2020pddlstream,garrett2018stripstream} attempts to increase the universality of PDDL by incorporating \textit{streams} into PDDL operators in addition to actions. A PDDLStream task can be represented as a tuple $T_{stream}=\langle Objects,S,G,Actions,Streams \rangle$. $Objects$ is a finite set of objects associated to the task, e.g., \texttt{body1} and \texttt{region2}. $S$ and $G$ are the initial and goal state of planning task respectively, each of which consists a set of literals. For the operators, $Actions$ contain all actions that can be used to update the environment; $Streams$ are operators that give out new objects as black-box generators of any kind. 
A skeleton $\pi_s$ composed of both actions and streams can be found as
\begin{align}\label{eq:5}
    &\langle \texttt{Sample-pose}(\texttt{body1},\texttt{region2}) \to \texttt{\#poseA},\nonumber\\ &\texttt{Plan-motion}(\texttt{body1},\texttt{pose1},\texttt{\#poseA)} \to \texttt{\#trajA}, \nonumber\\
    &\texttt{Pick-place}(\texttt{body1}, \texttt{region1}, \texttt{region2}, \texttt{\#trajA)} \rangle.
\end{align}        
Objects marked by $\#$ (e.g., \texttt{\#poseA}) are generated by streams and they demand \textit{bindings} of concrete values during binding search. 
These objects are described as \textit{optimistic} in the sense that they may not ever find their bindings under the geometric constraints.

\noindent\textit{\underline{Optimistic Expansion of Objects.}} In the PDDLstream description,
optimistic objects can be generated either at one stroke in the beginning or in incrementally across search rollouts \citep{garrett2018stripstream}. Here we take the first approach and summarize this optimistic enrichment procedure as $\#Objects,\#S=\text{OPTMS-EXPD}(Objects,S,Streams,level)$, as illustrated in Fig.\,\ref{fig:enrich}. $\#Objects$ and $\#S$ are the enriched objects and state respectively by applying $Streams$. For its detailed implementation we refer to \cite{garrett2018stripstream}.

\section{Top-k skeleton planning}\label{sec:topk-sk}

In task planning, we generate diverse skeletons via top-k planning.
Let $P_T$ be the set of all symbolic plans (possibly infinite) for a planning task $T$. The objective of top-k planning is to determine a set of \textit{k} different plans $\{\pi_1,\pi_2,...,\pi_k\}=P \subseteq P_T $ with the lowest costs for a given planning task \citep{speck2020symbolic}. We call a top-k algorithm \textit{complete} iff it could find the set of all plans as required. A standard top-k planner can be described as $P=\text{TOP-K}(Objects,S,G,Operators,k)$.

\subsection{Search for top-k action plans}
The planner firstly searches for the top-k action plans $P_a$ based on the optimistically enriched objects $\#Objects$ and initial state $\#S$. Those action plans are composed purely of actions. 
\begin{equation}
    P_a=\text{TOP-K}(\#Objects,\#S,G,Actions,k)
\label{eq:top-k}
\end{equation}

\subsection{Search for top-k skeletons}
Then the streams that generate those involved optimistic objects $\#Objects \setminus Objects$ are retracted and integrated to the action plans. Finally we have the top-k skeletons $P_s$ which are composed of both actions and streams. Each of these skeletons only relies on the original objects $Objects$ and initial state $S$.

\begin{equation}
\begin{aligned}
    \pi_s=\text{TOP-K}(Objects,S,G_s(\pi_a,G),
    Actions \cup Streams,1)
\end{aligned}
\label{eq:stream_retrace}
\end{equation}

$Actions$ and $Streams$ constitute the operator set of this planning problem. Informally, by setting $k=1$, the top-k algorithm reduces to a classical PDDL planner.
To ensure that the sequences of actions maintain from $\pi_a$ to $\pi_s$, we add extra constraints to the goal state. By assuming $\pi_a=\langle a_1,a_2,...,a_n \rangle$, we define the updated goal state $G_s$ as below.
$(a_1\prec a_2)$ is a literal asserting that $a_1$ is the predecessor of $a_2$.

\begin{equation}
    G_s(\pi_a,G)=\{(a_1\prec a_2),...,(a_{n-1}\prec a_n)\} \cup G,
\label{eq:10}
\end{equation}
 
\subsection{Algorithm summary}

\begin{algorithm}
\DontPrintSemicolon
\KwInput{$Objs,S,G,Actions,Streams,k$}
 \For{$level \in [0,1,...,max\text{-}level]$}
 {
  $\#Objs,\#S=\text{OPTMS-EXPD}(Objs,S,Streams,level)$, \\
  $P_a=\text{TOP-K}(\#Objs,\#S,G,Actions,k)$, \quad see \eqref{eq:top-k}\\
  \If{$|P_a|<k$}
  {
  \textbf{continue for}
  }
  $P_s=\{\,\}$\\
  \For{$\pi_a \in P_a$}
  {
    $Operators=Actions \cup Streams$\\
    $\#G=G_s(\pi_a,G)$, \quad see \eqref{eq:10}\\
    $\pi_s=\text{TOP-K}(Objs,S,\#G,Operators,1)$, see \eqref{eq:stream_retrace}\\
    $P_s \leftarrow \{\pi_s\} \cup P_s$
  }
  \textbf{return} $P_s$
 }
 \caption{\small TOP-K-SKELETON} \label{alg:topk-skeleton}
\end{algorithm}

An overview of the above skeleton planning algorithm is shown
in Alg.\,\ref{alg:topk-skeleton}. The inputs of the algorithm include the
original initial state $S$, goal state $G$, available $Streams$ and $Actions$, and a desired number of skeletons $k$. The output $P_s$ is the list of the top $k$ skeletons.
The size of $\#Objects$ ramps up quickly with increasing $level$. To address this, in Alg.\,\ref{alg:topk-skeleton} we regulate the number of optimistic objects by progressively increasing $level$. 
Here we choose the SYM-K algorithm as our top-k planning subroutines in \eqref{eq:top-k} and \eqref{eq:stream_retrace} due to its proved completeness and soundness  \citep{speck2020symbolic}. Accordingly we can derive the completeness property of our top-k skeleton planning algorithm. 

\begin{property} Probabilistic completeness in the top-k skeleton planning.\label{property:1}
    Given a complete top-k planner, any potentially feasible skeleton will be contained by the result set of Alg.\,\ref{alg:topk-skeleton} as $k$ goes to infinity. 
\end{property}

After applying Alg.\,\ref{alg:topk-skeleton} to the toy planning problem task (Fig.\,\ref{fig:toy_problem}), we get a feasible skeleton as shown in Table\,\ref{tab:skeleton_layer}.

\begin{table}
    \caption{\small An example skeleton to the toy problem in Fig.\,\ref{fig:toy_problem} with its operators grouped by tree node layers.} \label{tab:skeleton_layer}
\centering
\begin{tabular}{c|c}
    \hline
    \makecell{Layers} & \makecell{Operators}\\
    \hline
    \textbf{decision1} & \makecell[l]{$\texttt{Sample-pose(body2,region1)}$
    $\to \texttt{\#poseB}$}\\
    \hline
    \textbf{transition1} & \makecell[l]{$\texttt{Plan-motion(body2,pose2,\#poseB)}$ 
    $\to \texttt{\#trajB}$ \\
    $\texttt{Pick-place(body2,region1,region1,\#trajB)}$} \\
    \hline
    \textbf{decision2} & \makecell[l]{$\texttt{Sample-pose(body1,region2)}$ 
    $\to \texttt{\#poseC}$} \\
    \hline
    \textbf{transition2} & \makecell[l]{$\texttt{Plan-motion(body1,pose1,\#poseC)}$ 
    $\to \texttt{\#trajC}$ \\
    $\texttt{Pick-place(body1,region1,region2,\#trajC)}$} \\
    \hline
    \end{tabular}
    \vspace{-0.5cm}
\end{table}

\section{Tree search in the extended decision space}\label{sec:extended-tree-search}
Based on the top-k skeletons generated above, we must search for feasible concrete plans that are geometrically feasible and further find the optimal plan with the highest reward. To achieve this, firstly, we have to choose from the candidate skeletons (\textit{skeleton space}) the one that we would like to continue with in binding search (Sec.\,\ref{subsec:skeleton_selection}); secondly, we must search for concrete bindings of its symbolic variables (e.g., \texttt{\#pose12} in \eqref{eq:5}) in a \textit{motion space} (Sec.\,\ref{subsec:binding_search}). In this study, we define a reward function  \eqref{eq:reward_fun} as the optimization target for both skeleton selection and binding search. For a skeleton $\pi_i$ with $H_i$ symbolic variables to bind (\textit{binding horizon}), we have 

\begin{equation}
r = p_t\left(\frac{1+d_\text{end}}{H_{i}}+\frac{p_m}{motionCost+1}\right)+r_\text{end}.
\label{eq:reward_fun}
\end{equation}

$d_{end}$ is the number of symbolic variables that have already found their feasible bindings, and $motionCost$ is proportional to the swept volume of the robot from its initial state to the terminal state. The first term in \eqref{eq:reward_fun} encourages the planner to avoid choices where less bindings are found. The second term makes the planner prefer choices with motions that have less occupation in the workspace. For the third term, $r_\text{end}=1$ when all bindings are successfully found, otherwise $r_\text{end}=0$. For the hyper parameters, we set $p_t=0.1$ and $p_m=1.0$.

\subsection{Optimal search in the binding space} \label{subsec:binding_search}

A skeleton describes logically a path
to the goal state. However, its geometric feasibility must be evaluated
along with the sequential binding of its symbolic variables to
concrete values. We describe the binding search problem as a Markov Decision Process (MDP) and address it as a classical finite-horizon stochastic optimal search problem. 

We propose using PW-UCT to systematically explore this decision space. Since we have a mixtured motion space with both discrete and continuous decisions, we employ a UCT variant with progressive widening (PW) techniques \citep{coulom2007computing,auger2013continuous,chaslot2008progressive}. Its basic idea is to limit the number of visits for existing nodes artificially. When the value of the existing nodes is estimated sufficiently well, new nodes will be expanded to explore the unvisited regions of the decision spaces. At a decision node $z$, if $\text{PW-TEST}(z)$ is true, a new child (or binding decision) of $z$ will be sampled and thereby the tree will grow a new branch; otherwise, one of its existing children will be selected by UCB.

\begin{equation}
\text{PW-TEST}(z)=\floor{\text{visits}(z)^\alpha}>\floor{(\text{visits}(z)-1)^\alpha}
\label{eq:expandable-by-pw}
\end{equation}

In \eqref{eq:expandable-by-pw} index $\alpha$ is the corresponding PW constant for balancing expansion with evaluation \cite{auger2013continuous}. With a large $\alpha$, the planner will expand eagerly its skeleton space; when $\alpha$ is small, the planner will be reluctant to take new skeletons and will spend more time evaluating the existing ones.

The search tree of PW-UCT is structured with interleaved layers of decision nodes and transition nodes. We group our skeleton by these two node types as illustrated in Table.\,\ref{tab:skeleton_layer}. Decision nodes are responsible for the independent bindings (e.g., \texttt{\#pose21a} in \textbf{decision1}); while transition nodes decides on the dependent bindings (e.g., \texttt{\#traj211a} in \textbf{transition1}) and update the environment state (e.g., \texttt{Pick-place} from \textbf{transition1}). Since transition nodes contain sample-based generators (e.g., RRT \cite{lavalle1998rapidly} in \texttt{Plan-motion}), their resulting states can be different (see \textbf{state2.2} and \textbf{state2.3} in Fig.\,\ref{fig:tree}) even with the same starting state and the same binding decisions.

\subsection{Optimal search in the skeleton space} \label{subsec:skeleton_selection}

We model skeleton selection as a multi-armed bandit problem with the reward function of \eqref{eq:reward_fun}, and solve it via UCB1~\citep{auer2002finite}. With a UCB criterion, we are able to maintain an optimal balance between exploring new skeletons and exploiting existing skeletons with high rewards. The regrets of each skeleton selection are therefore kept to minimum. 

Before any binding search, we have $d_{end}=0$, $r_\text{end}=0$, and $motionCost=0$. In this case, the bandit agent would prefer skeletons with small binding horizons. It is a reasonable behaviour in the absence of other heuristics, since less bindings  indicates not only less chance of failing in the binding search, but also less actions included in the skeleton. When some binding search results are available, this reward function takes more factors into consideration when evaluating a skeleton. It will prioritize skeletons that have been proved geometrically feasible (the third term), skeletons with high percentage of bindings found (the first term), and skeletons whose motion cost is low (the second term).

To ensure complete search in task planning, we should have $k$ in TOP-K-SKELETON in Alg.\,\ref{alg:topk-skeleton} sufficiently large.
When $k \rightarrow \infty$, a continuum-armed bandit instead of a multi-arm one should be used to model the extended root. Inspired by the solution to the continuous motion space, we use the progressive widening trick again at the decision node (noted as $e\_root$) for skeleton selection. As shown in \eqref{eq:expandable-by-pw}, if $\text{PW-TEST}(e\_root)$ is true, then a new skeleton is added to the skeleton space $P_s$, otherwise the bandit will continue searching the existing skeletons. 

\subsection{The extended decision tree}

\begin{figure}
    \centering
    \includegraphics[width=0.46\textwidth]{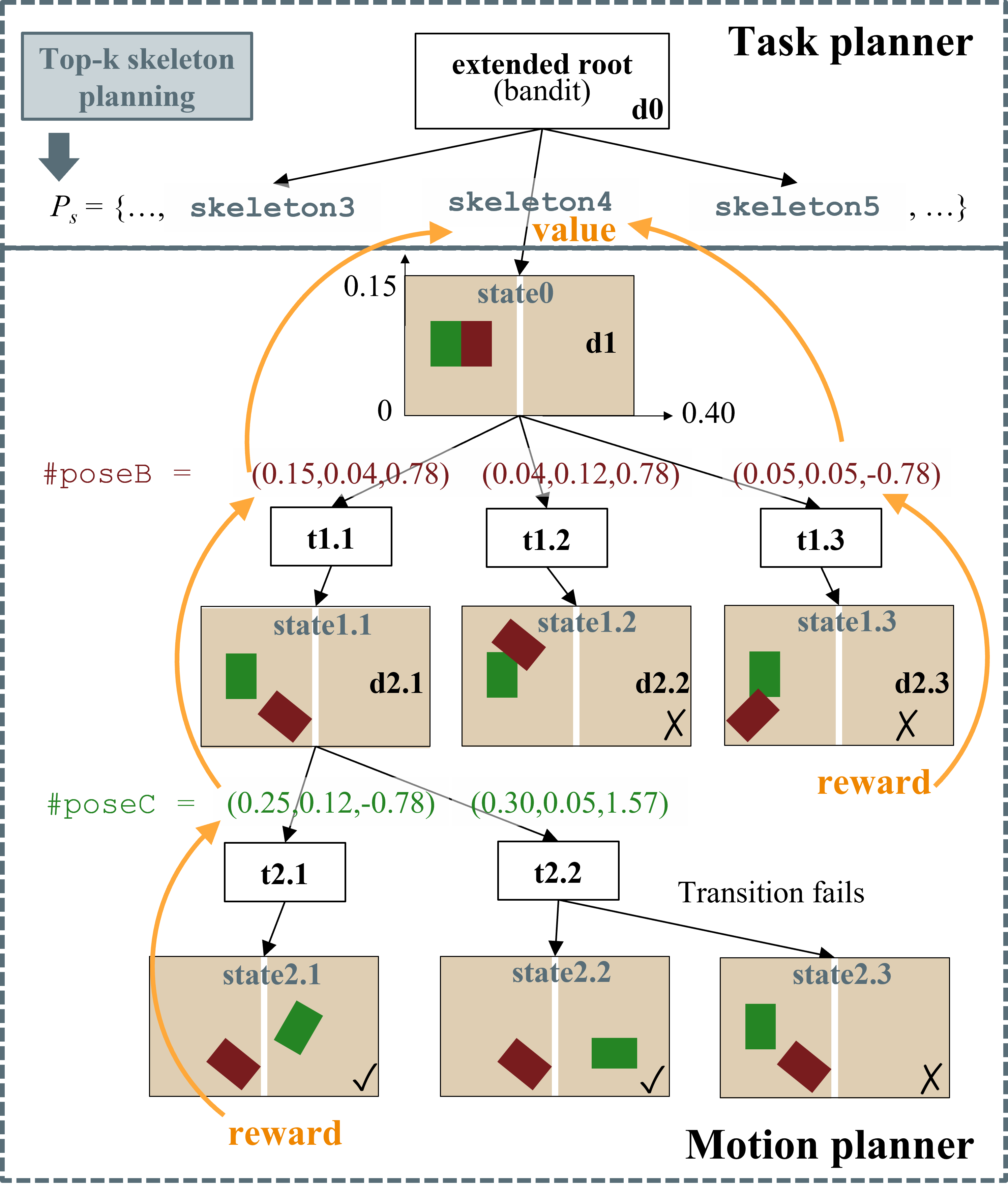}
    \caption{\small The extended decision tree: an integration of optimal skeleton selection and binding search.}
    \label{fig:tree}
    \vspace{-0.5cm}
\end{figure}

Based on a given skeleton, we can build a decision tree for binding search as shown in the lower part of Fig.\,\ref{fig:tree}. The motion planner take a skeleton as the input and search for its bindings by PW-UCT. At each decision node (e.g., \textbf{d1}), the planner samples a binding for a symbolic variable (e.g., \texttt{\#pose21a}) and update the simulator correspondingly through a transition node (e.g., \textbf{t1.1}). If the resulting state is geometrically feasible (e.g., \textbf{state1.1}), the planner is able to continue binding the next symbolic variable (e.g., \texttt{\#pose12a}). Otherwise, a bad binding decision leads to a geometrically infeasible state (e.g., \textbf{state1.2}) and the current branch must terminate here.
Whenever a terminate node is created, it propagates its reward through its parents all the way back to the root of the current skeleton branch  \cite{coulom2006efficient}. 

In this way, the value of selecting of that specific skeleton (e.g., \texttt{skeleton4}) is updated (orange arrows in Fig.\,\ref{fig:tree}) and it can be used by the multi-arm bandit in skeleton space (the upper part in Fig.\,\ref{fig:tree}). Further observation of Fig.\,\ref{fig:tree} reveals that the skeleton selection process is no more than an extended decision node before bind search and it can be readily included into the UCT structure. By having the bandit of skeleton selection as an extended root, we define an extended decision tree that is able to search in both the skeleton space and the motion space. 

\subsection{Algorithm summary}

We can summarize the extended tree search algorithm in Alg.\,\ref{alg:tree_search}. It largely remains the structure of a standard PW-UCT algorithm. The children of the extended root node $e\_root$ that represents the skeleton space is initialized with empty set.

\begin{algorithm}
\DontPrintSemicolon
\KwInput{$P_s,t_{ts}$}
$new\_skeletons=P_s-e\_root.\text{CHILDREN}$\\
$e\_root=e\_root.\text{ADD-CHILDREN}(new\_skeletons)$\\
$node \leftarrow e\_root$\\
\While{$timeCost()<t_{ts}$}
{
    \While{$node$ is not terminated}
    {
        ADD-VISIT($node$)\\
        \eIf{PW-TEST$(node)$}
        {
            $node \leftarrow \text{EXPAND-NEW-CHILD}(node)$, 
        }
        {
            $node \leftarrow \text{SELECT-CHILD-BY-UCB}(node)$, 
        }
    }
    $\text{BACK-RPOPAGATE}(node)$\\
    \vspace{1.0mm}
    $node \leftarrow e\_root$\\
}
$\pi_c=e\_root.\text{BEST-CHILD}$\\
\textbf{return} $\pi_c$
\caption{\small EXTENDED-TREE-SEARCH}\label{alg:tree_search}
\end{algorithm}

The inputs of Alg.\,\ref{alg:tree_search} include the top-k skeletons $P_s$ from Alg.\,\ref{alg:topk-skeleton} as new children of the extended root, and a user-defined time budget $t_{ts}$. Its output is a single concrete plan $\pi_c$ that is the best child of the extended root in terms of branch value.
On ADD-VISIT$(node)$, $node$ will increase its visit number by 1. Meanwhile, the environment will be set to the initial state of $node$. By complying with UCB and PW criterion, a consistent value estimation of each node (e.g., \textbf{d1} in Fig.\,\ref{fig:tree}) is guaranteed. 

\begin{property} Consistent value estimation in extend tree search.\label{property:2}
     The value estimation of each tree node will convergence to its real value as the search time goes infinite.
\end{property}

The above property can be perceived  straightforwardly via a recursive analysis from the leaf nodes to the root, we refer to \citep{auger2013continuous} for the detailed proof. 
Based on the consistent estimation of decision value, a global convergence of the output $\pi_c$ to the optimal plan is expected. 

\begin{property}Probabilistic completeness of extended tree search.\label{property:3}
     The optimal concrete plan within the given skeleton space (implied by $P_s$) can be found after $t_{ts}$ goes infinite.
\end{property}

\section{The eTAMP algorithm}\label{sec:etamp}
We introduce some techniques to make the proposed method more efficient in practice (Sec.\,\ref{subsec:incremetal_skeleton} and Sec.\,\ref{subsec:delay_binding}), and summarize the final algorithm (Sec.\,\ref{subsec:etamp}).

\subsection{Incremental skeleton space} \label{subsec:incremetal_skeleton}

In theory, a TAMP framework as presented in Fig.\,\ref{fig:tree} is capable to explore the all combinations of skeletons and their bindings when the skeleton space $P_s$ is exhaustive. However, we have found that the even for state-of-the-art top-k planners like sym-k, a large $k$ normally means significantly elongated planning time. To balance efficiency with completeness in task planning, we propose to incrementally expand the skeleton during search. We divide the whole TAMP into sessions with incremental skeletons. 
In each session, a batch of new skeletons is added to the skeleton space. The TAMP system searches the current skeleton space for a given time period and then comes to the next session.

\subsection{Delayed binding of of new skeletons} \label{subsec:delay_binding}

By enabling knowledge sharing across skeleton branches, we can improve the overall planning efficiency.
Although the candidate skeletons are different from each other, they may share the same operator sequences at their initial parts. For example in Fig.\,\ref{fig:candidate_sk}, \texttt{skeleton1} and \texttt{skeleton3} have the same first high-level action \texttt{action1}. If the binding of \texttt{action1} in \texttt{skeleton1} is not found at the moment, there is no reason to start binding it in \texttt{skeleton3}. The binding of \texttt{skeleton3} should be delayed in this case so that the skeleton space can be kept as small as possible.
On the other hand, whenever a feasible binding is found for this operator in \texttt{skeleton1}, it should be shared with \texttt{skeleton3} and thus gives the binding search of \texttt{skeleton3} a warm start.

\subsection{Algorithm summary} \label{subsec:etamp}

\begin{algorithm}[h!]
\DontPrintSemicolon
\KwInput{$T_{stream},k_{batch},t_{ts},t_{d}$}
$i=1$\\
$P_s=\{\,\}$\\
$e\_root.\text{CHILDREN}=\{\,\}$\\
\While{$timeCost()<t_{d}$}
{
$k=k_{batch}\cdot i$\\
$P_s=P_s\,|\,\text{TOP-SKELETON}(T_{stream}, k)$, see Alg.\,\ref{alg:topk-skeleton}\\
$\pi_c=\text{EXTENDED-TREE-SEARCH}(P_s,t_{ts})$, see Alg.\,\ref{alg:tree_search}\\
$i=i+1$\\
}
\textbf{return} $\pi_c$
\caption{\small eTAMP}\label{alg:etamp}
\end{algorithm}

A serial combination of Alg.\,\ref{alg:topk-skeleton} and Alg.\,\ref{alg:tree_search} gives out the overall eTAMP algorithm, as summarized in Alg.\,\ref{alg:etamp}. It takes as input a PDDLStream problem description $T_{stream}=\langle Objects,S,G,Actions,Streams \rangle$, and three hyper-parameters: $k_{batch}$, $t_{ts}$ and $t_b$. $k_{batch}$ is the number of new skeletons added to the skeleton space in the next extended tree search session, and $t_{ts}$ is the time cost allocated to each session. $t_b$ limits the total planning time of eTAMP.
Given sufficiently large $t_b$, $k$ in Alg.\,\ref{alg:etamp} will grow to infinity, and the extended root will have all possible skeletons as it children (\textit{property \ref{property:1}}).
Meanwhile, as the visit number of the extended root increases, the resultant concrete plan $\pi_c$ will converge to the optimal one with respect to the reward function \eqref{eq:reward_fun} (\textit{property \ref{property:3}}). 

\section{Empirical evaluation} 
\label{sec:evaluation}

We empirically evaluate the proposed eTAMP algorithm in four domains: Kitchen, Hanoi Tower, Unpacking and Regrasping. Wherein Kitchen and Hanoi Tower are adapted from the framework-independent benchmarks for TAMP system proposed by \cite{lagriffoul2018platform}, and the other two are based on the existing robot hardware of our lab. We list the recognizable difficulties that  each task represents for testing the planner in Table\,\ref{tab:4}. In addition to those criteria listed in \cite{lagriffoul2018platform}, we add another dimension \textit{large motion spaces} to measure the difficulty in binding search. This criterion will be satisfied if the underlying motion planning problem of the domain requires substantial search effort. It applies to tasks with many bindings to be decided or some of their bindings have tight feasible spaces.

\begin{table}
    \caption{\small The difficulties reflected by the evaluation domains} \label{tab:4}
\centering
\begin{tabular}{c|c|c|c|c}
    \hline
    \makecell{Difficulty} & \makecell{Kitchen} & \makecell{Tower} & \makecell{Unpacking} & \makecell{Regrasping} \\
    \hline
    Infeasible task actions &  &  & \checkmark & \checkmark \\
    \hline
    Large task space &  & \checkmark &  &  \\
    \hline
    Large motion space* & \checkmark & \checkmark &  & \checkmark \\
    \hline
    Non-monotonicity &  &  &  & \checkmark\\
    \hline
    Non-geometric actions & \checkmark &  &  & \checkmark\\
    \hline
    \end{tabular}
    *An extra criterion in addition to those of \cite{lagriffoul2018platform}.
    \vspace{0.0cm}
\end{table}

 The Adaptive algorithm from \cite{garrett2020pddlstream} represents the best performance of existing PDDLStream methods, and it serves as a baseline in evluating eTAMP. The two algorithms share one internal simulator that is implemented with PyBullet \cite{coumanspybullet} for hosting the environment state. Both planners are terminated after a 300-second timeout for each of the three tasks. For eTAMP we set $k_{batch}=50, t_{ts}=250$. In the extened tree search (see Alg.\,\ref{alg:tree_search}),  The MCTS rollout policy is random and no heuristics is used to guide the search. PW-TEST (see \eqref{eq:expandable-by-pw}) takes $\alpha=0.26$ for regulating the growth of new branches. The performance of each algorithm is evaluated over 100 instances for each task. Whenever a feasible concrete plan is found, the algorithm is stopped and the computation time is calculated. These instances are diversely generated by randomly initializing the initial geometric state. It is noteworthy that both algorithms employ the same set of random stream generators for motion bindings, which have zero heuristics. 
 
 \begin{figure}
    \centering
    \includegraphics[width=0.49\textwidth]{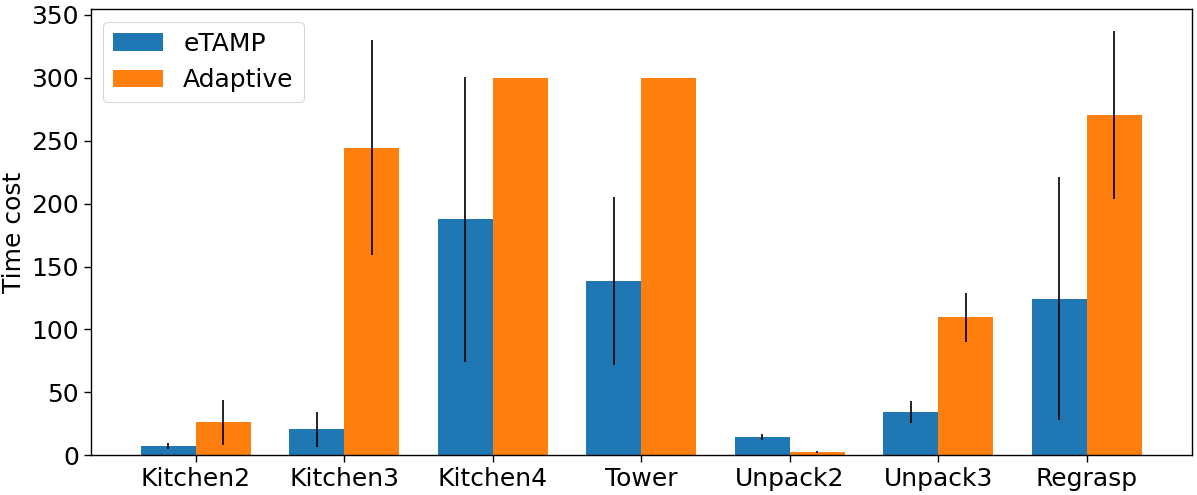}    \vspace{-0.5cm}
    \caption{\small The average time cost with std indicators of the TAMP methods in solving the evaluation tasks.}
    \label{fig:results}
    \vspace{-0.5cm}
\end{figure}

\subsection{Kitchen domain} 

\begin{figure}
    \centering
    \includegraphics[width=0.49\textwidth]{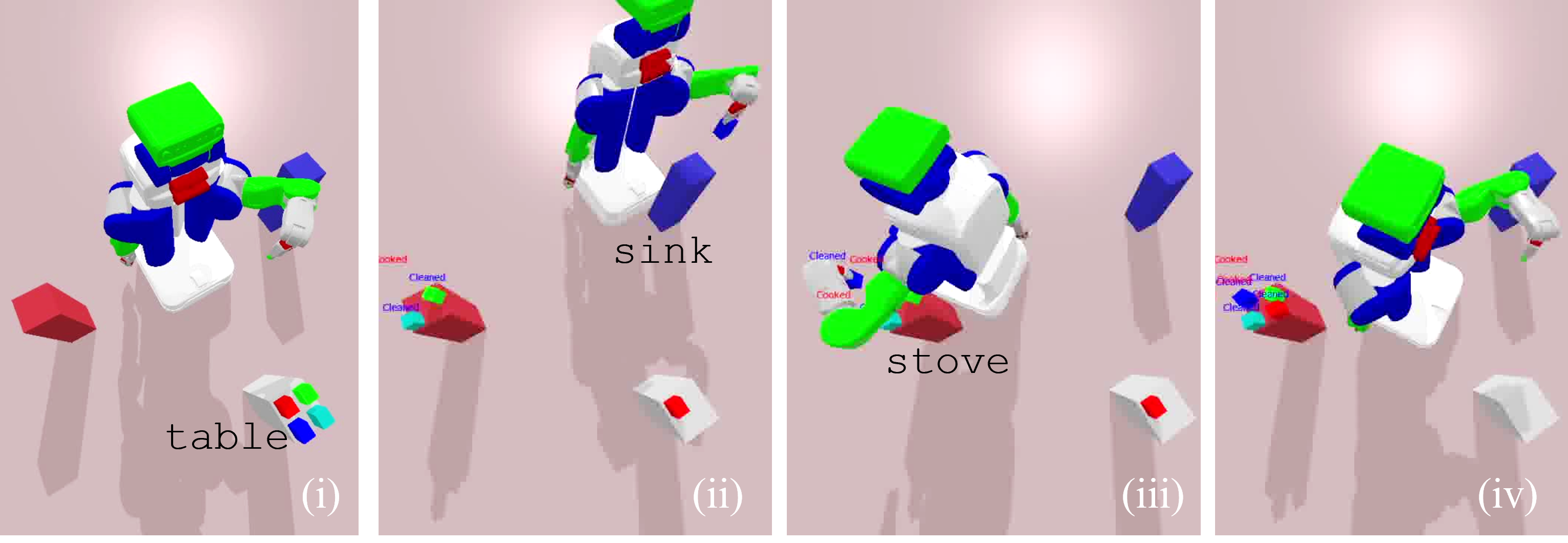}    \vspace{-0.5cm}
    \caption{\small Kitchen domain. Given an initial number of bodies to be  cooked, the mobile manipulator should cook by first placing the body on the sink for cleaning, and then place it on the stove for cooking.}
    \label{fig:kitchen_domain}
    \vspace{-0.5cm}
\end{figure}

\begin{figure}
    \centering
    \includegraphics[width=0.49\textwidth]{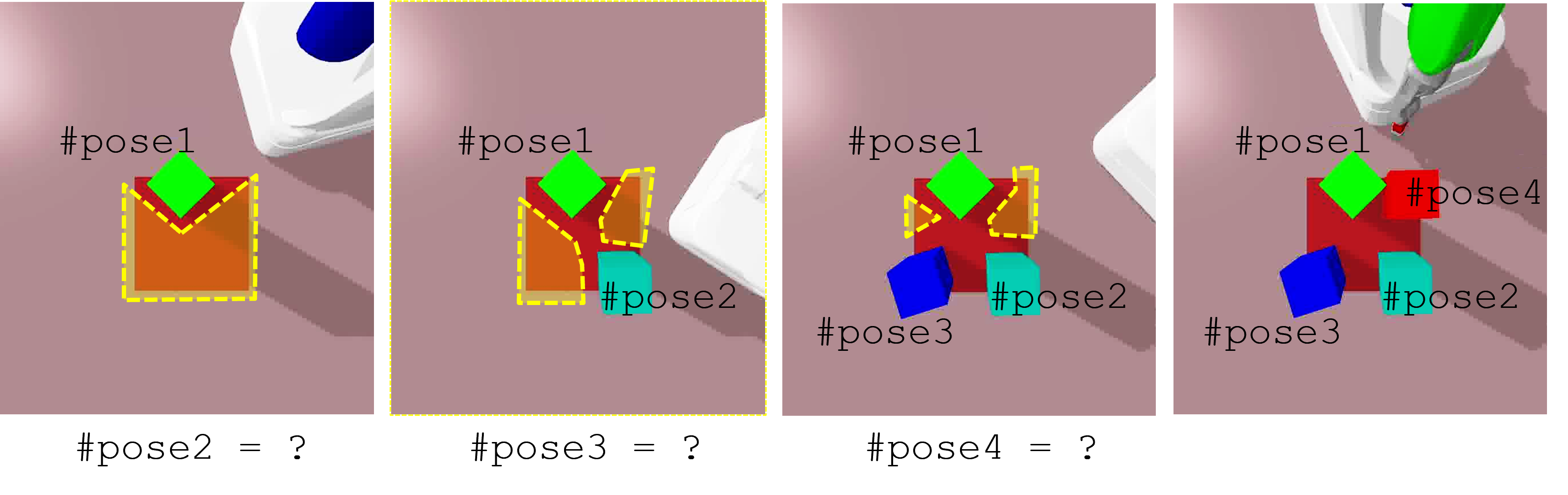}    \vspace{-0.5cm}
    \caption{\small The \textit{large motion space} of the Kitchen domain. The feasible space (the yellow zone) for the pose of coming bodies shrinks significantly by each placement. Binding search for these poses is challenging given zero heuristics.}
    \label{fig:kitchen_difficulty}
    \vspace{-0.5cm}
\end{figure}

Compared to the original Kitchen task in \cite{lagriffoul2018platform}, this domain reduces the complexity in task planning but meanwhile becomes more demanding to the motion planner. As shown in Fig.\,\ref{fig:kitchen_domain}, the mobile manipulator must "cook" a given number of food blocks initially placed on \texttt{table}. Food must be cleaned before it can be cooked. It can be cleaned when placed on \texttt{sink} and cooked when placed on \texttt{stove}. The metric decisions left to the motion planner is the body pose of each placement. The actions of \texttt{Clean} (via \texttt{sink}) and \texttt{Cook} (via \texttt{stove}) are non-monotonic from the robot’s perspective. The main difficulty of this task lies in the tight stove region, where several bodies need to be placed. The search of the feasible pose for the newly coming body becomes exponentially challenging after each placement. In this way, this domain evaluates the planners' performance in \textit{large motion space}. As the planners are not provided with this potential congestion in their symbolic knowledge, they must search for the feasible decisions through trial and error. During binding search, the consistent value estimation of each decision is critical for this delayed-reward case. 

We show the evaluation results in Fig.\,\ref{fig:results} where Kitchen2, Kitchen3, Kitchen4 indicate three different settings with the number of bodies to be cooked equals to 2, 3, 4, respectively. eTAMP scales better than Adaptive to more complicated settings and even has some chance solving the problem in Kitchen4. More specifically, in terms of the successful rates, both methods achieved 100\% for Kitchen2. eTAMP achieved 100\% and 60\% for Kitchen3, Kitchen4, respectively, while Adaptive achieved 36\% and 0\%.

\subsection{Hanoi Tower domain} 

\begin{figure}
    \centering
    \includegraphics[width=0.49\textwidth]{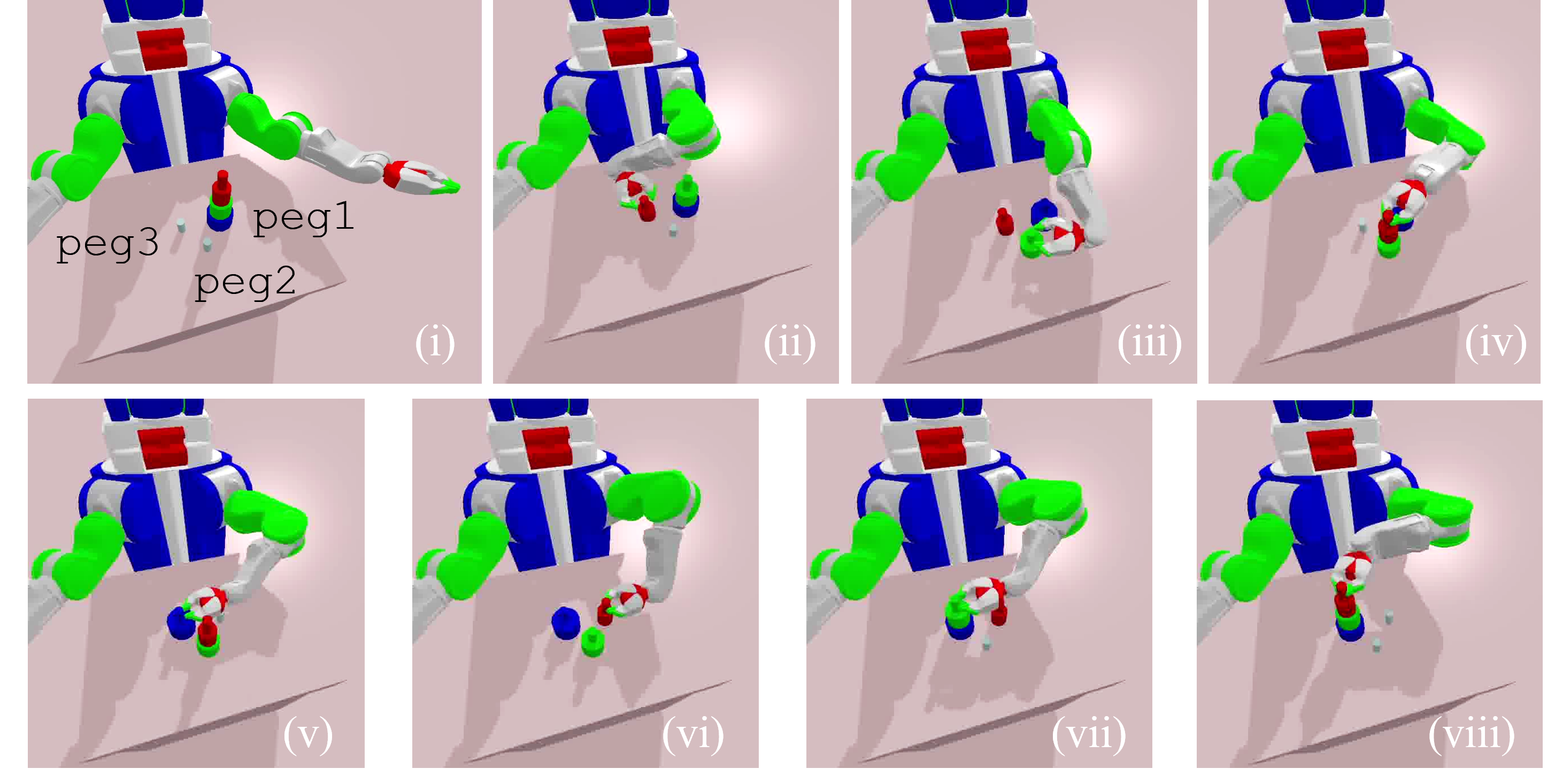}
    \caption{\small Hanoi Tower domain. The robot must move all discs from \texttt{peg1} to \texttt{peg3} without causing any collision in motion spaces while complying to the game rule in task spaces.}
    \label{fig:hanoi_domain}
    \vspace{-0.5cm}
\end{figure}

\begin{figure}
    \centering
    \includegraphics[width=0.49\textwidth]{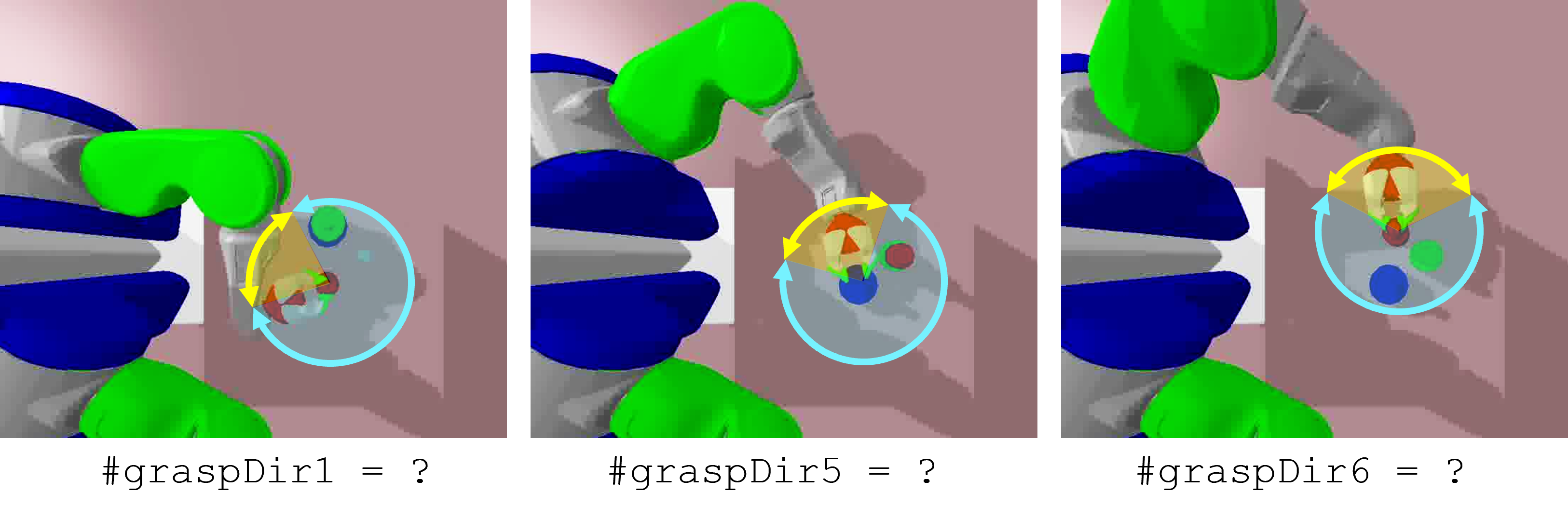}
    \vspace{-0.5cm}
    \caption{\small The \textit{large motion space} of the Hanoi Tower domain. The feasible grasp direction (the yellow zone) is small compared to the whole decision space. Binding search for these poses is challenging given zero heuristics. The motion planner has to bind \texttt{\#graspDir1},\texttt{\#graspDir2},...,\texttt{\#graspDir8} to ground values (degrees) for the task completion.}
    \label{fig:hanoi_difficulty}
\end{figure}

This domain remains mostly same to its original version expect that here only one arm of the PR2 robot is allowed to manipulate (Fig.\,\ref{fig:hanoi_domain}). A stack of three discs of different sizes must be relocated to a target pole via a intermediate pole while obeying the game rule that the upper disc should always be smaller than the lower one. After generating a skeleton in the \textit{large task space}, the motion planner is tasked to do binding search for the grasping direction of each pick-place operation. Due to the confined work space and joint limitations, the robot must search carefully this \textit{large motion space}. 

As shown in Fig.\,\ref{fig:results}, eTAMP got a successful rate of 99\% in the evaluation while Adaptive got 0\% and could not give a solution within the time limitation. We attribute the poor performance of Adaptive in this domain to the tight decision space in binding search. As shown in Fig.\,, the feasible binding area is small compared to the large search space ($[-\pi,\pi]$ in this example) due to the obstruction of nearby entities and the kinematic configuration of the robot arm. The robot has to make eight successive decisions like this to complete the task. Thus this domain is quite challenging to TAMP methods that are not dedicated to optimize the exploitation-exploration behaviour in motion planning such as Adaptive. 

\subsection{Unpacking domain} 

\begin{figure}
    \centering
    \includegraphics[width=0.45\textwidth]{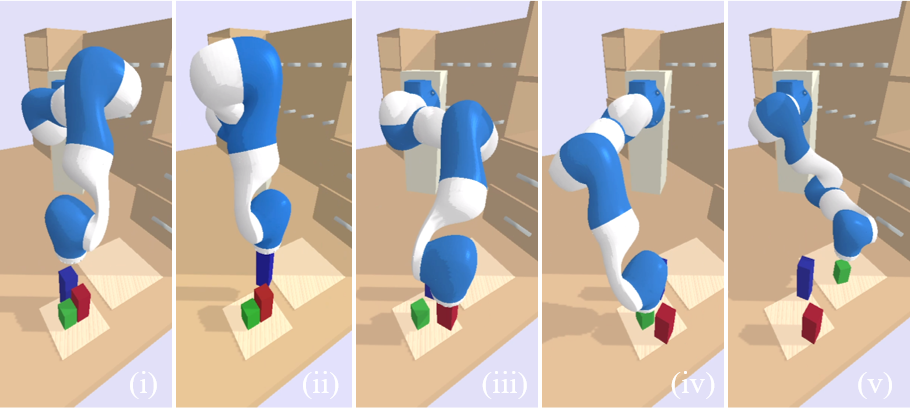}
    \caption{\small The 3-body Unpacking task. To reach the green body, taller bodies must be relocated by the robot at first.}
    \label{fig:unpacking}
\end{figure}

This task comes directly from the motivation example of Fig.\,\ref{fig:1}. When more than one body is involved in the scene, the problem becomes challenging since the task planners are not informed of the geometric constraints, as imposed by the taller red body in Fig.\,\ref{fig:toy_problem} (\textit{infeasible task actions}). In addition the 2-body scenario (Unpack2), we also tested the TAMP algorithms in a 3-body scenario (Unpack3) as shown in Fig.\,\ref{fig:unpacking}, where two taller bodies must be relocated before the target one can be reached. 

Both methods solved Unpack2 and Unpack3 with a successful rate of 100\%. However, when look into the details in Fig.\,\ref{fig:results}, the two methods showed different characters. When motion planning is simple as the Unpack2 setting, eTAMP cost more time than Adaptive since it has to generate the top-k skeletons at the first place while most of them are unnecessary in this case. However, in Unpack3, when the the motion planning space gets larger (though not as large as the the other evaluation domains), eTAMP gave better performance due to its more efficient binding search.

\subsection{Regrasping domain} 

\begin{figure}
    \centering
    \includegraphics[width=0.49\textwidth]{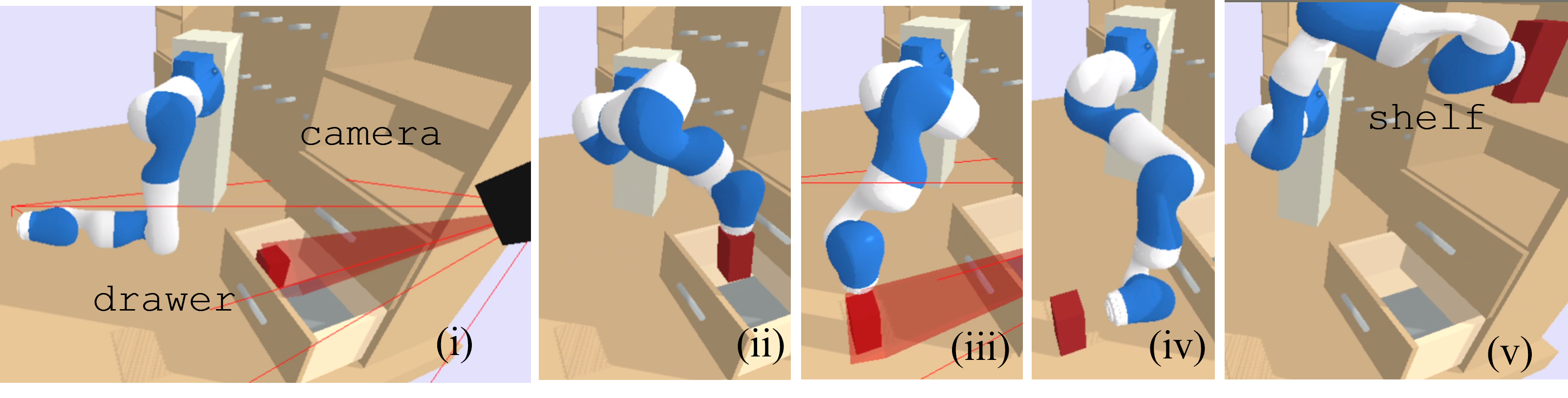}
    \caption{\small Regrasping domain. The only way the robot could move the cuboid body from the drawer to the shelf is by regrasping it with a different direction after taking it from the drawer.}
    \label{fig:regrasp_domain}
    \vspace{-0.5cm}
\end{figure}

As illustrated in Fig.\,\ref{fig:regrasp_domain}, the robot must transport the red block from the drawer to the shelf, where it should still be placed upright. After each displacement, the location of the block must be measured by a camera. We have \texttt{locate-object} as a \textit{non-geometric action}. Due to the geometric constraints of the environment, the robot cannot move the block to its target area directly without regrasping it from a different direction. 
The robot can choose from 5 directions to grasp the block: along the normal vectors of the top surface and four side surfaces. The block at its initial pose is limited to top grasping by \texttt{drawer}, while it must be grasped from the side when being placed on \texttt{shelf}, as depicted in Fig.\,\ref{fig:regrasp_domain}. This domain is difficult for task planning since a regrasping behavior must be composed. Otherwise, the problem cannot be solved (\textit{infeasible task actions}). Moreover, this task is geometrically challenging. Even if a correct skeleton is given, the motion planner must search the hybrid decision space (discrete grasping directions and continuous body poses) for bindings (\textit{large motion space}).

Fig.\,\ref{fig:results} shows eTAMP in this challenging task performs better than Adaptive. eTAMP presented a successful rate of 86\% compared to that of 22\% from Adaptive.

\subsection{Summary} 
In the above experiments, we evaluate the time cost of the considered TAMP algorithms for finding the first feasible plans. Optimality (e.g., motion cost) of these plans is not considered in this study. 
Evaluation results show competitive performance of eTAMP in solving challenging TAMP problems as summarized in \cite{lagriffoul2018platform}. We have to admit that most state-of-the-art TAMP methods such as Adaptive is satisfying in solving tasks with \textit{infeasible task actions} and \textit{large task space}, as shown in the Unpack domain in Fig.\,\ref{fig:results}. On the other hand, the unique tree search in the extended decision space makes eTAMP especially efficient when the motion space gets large and existing binding search practices start to suffer, which is proved by the experiments with the Kitchen domain, the Hanoi Tower domain and the Regrasping domain.

\section{Conclusion} 
\label{sec:conclusion}

We propose eTAMP as a general-purpose planner of robot manipulation tasks with long horizons that demand symbolic sequencing of operators and binding search of motion parameters under geometric constraints. 
Similar to other general-purpose TAMP frameworks, eTAMP relies on its internal symbolic planner to deal with the \textit{large task space} and \textit{non-geometric actions}. What makes it unique is that it derives its capability of addressing the \textit{infeasible task actions} from the diverse backup skeletons generated by a top-k planner and an optimal selection process of those skeletons. By modeling the binding search as MDPs, eTAMP utilizes PW-UCT to find ground values for the symbolic variables in skeletons and thus it is more prepared for the \textit{large motion space} that is ubiquitous in realistic robotic manipulation.

\bibliography{references}

\end{document}